\documentclass[sigconf]{acmart}
\AtBeginDocument{%
  }

\setcopyright{acmcopyright}
\copyrightyear{2018}
\acmYear{2018}
\acmDOI{XXXXXXX.XXXXXXX}

\acmConference[*]{Make sure to enter the correct
  conference title from your rights confirmation emai}{2023}{}
\acmPrice{15.00}
\acmISBN{978-1-4503-XXXX-X/18/06}

\usepackage{subfigure}
\usepackage{multirow}
\usepackage{colortbl}
\usepackage[ruled,linesnumbered]{algorithm2e}
\usepackage{algorithmic}
\begin{document}

\title{Mover: Mask and Recovery based Facial Part Consistency Aware Method for Deepfake Video Detection}
\author{Juan Hu}

\affiliation{%
  \institution{College of Computer Science and Electronic Engineering, Hunan University, China}
  \streetaddress{}
  \city{}
  \state{}
  \country{}
  \postcode{}
}
\author{Xin Liao}
\affiliation{%
  \institution{College of Computer Science and Electronic Engineering, Hunan University, China}
  \streetaddress{}
  \city{}
  \state{}
  \country{}
  \postcode{}
}
\author{Difei Gao}
\affiliation{%
  \institution{Show Lab, National University of Singapore, Singapore}
  \streetaddress{}
  \city{}
  \state{}
  \country{}
  \postcode{}
}
\author{Satoshi Tsutsui}
\affiliation{%
  \institution{ROSE Lab, Nanyang Technological University, Singapore}
  \streetaddress{}
  \city{}
  \state{}
  \country{}
  \postcode{}
}
\author{Qian Wang}
\affiliation{%
  \institution{School of Cyber Science and Engineering, Wuhan University, China}
  \streetaddress{}
  \city{}
  \state{}
  \country{}
  \postcode{}
}
\author{Zheng Qin}
\affiliation{%
  \institution{College of Computer Science and Electronic Engineering, Hunan University, China}
  \streetaddress{}
  \city{}
  \state{}
  \country{}
  \postcode{}
}
\author{Mike Zheng Shou}
\affiliation{%
  \institution{Show Lab, National University of Singapore, Singapore}
  \streetaddress{}
  \city{}
  \state{}
  \country{}
  \postcode{}
}

\renewcommand{\shortauthors}{Hu et al.}

\begin{abstract}

Deepfake techniques have been widely used for malicious purposes, prompting extensive research interest in developing Deepfake detection methods. 
Deepfake manipulations typically involve tampering with  facial parts, which can result in inconsistencies across different parts of the face. For instance, Deepfake techniques may change smiling lips to an upset lip, while the eyes remain smiling. Existing detection methods depend on specific indicators of forgery, which tend to disappear as the forgery patterns are improved. 
To address the limitation,  we propose \textsf{Mover}, a new Deepfake detection model that exploits unspecific facial part inconsistencies, which are inevitable weaknesses of Deepfake videos. \textsf{Mover}  randomly \underline{m}asks  regions of interest (ROIs)  and rec\underline{over}s  faces to learn unspecific features, which makes it difficult for fake faces  to be recovered, while real faces can be easily recovered. 
Specifically, given a real face image, we first pretrain a masked autoencoder to learn facial part consistency by  dividing faces into three parts and randomly masking ROIs,  which are then recovered based on the unmasked facial parts. Furthermore, to maximize the discrepancy between real and fake videos, we propose a novel model with dual networks that utilize the pretrained encoder and masked autoencoder, respectively.
1) The pretrained encoder is finetuned for capturing the encoding of inconsistent information in the given video. 2) The pretrained masked autoencoder is utilized for mapping faces and distinguishing real and fake videos. Our extensive experiments on standard benchmarks demonstrate that \textsf{Mover} is highly effective. 
\end{abstract}

\renewcommand\footnotetextcopyrightpermission[1]{}
\settopmatter{printacmref=false} 


\maketitle



\begin{figure}[t]
  \centering
   \includegraphics[width=0.8\linewidth]{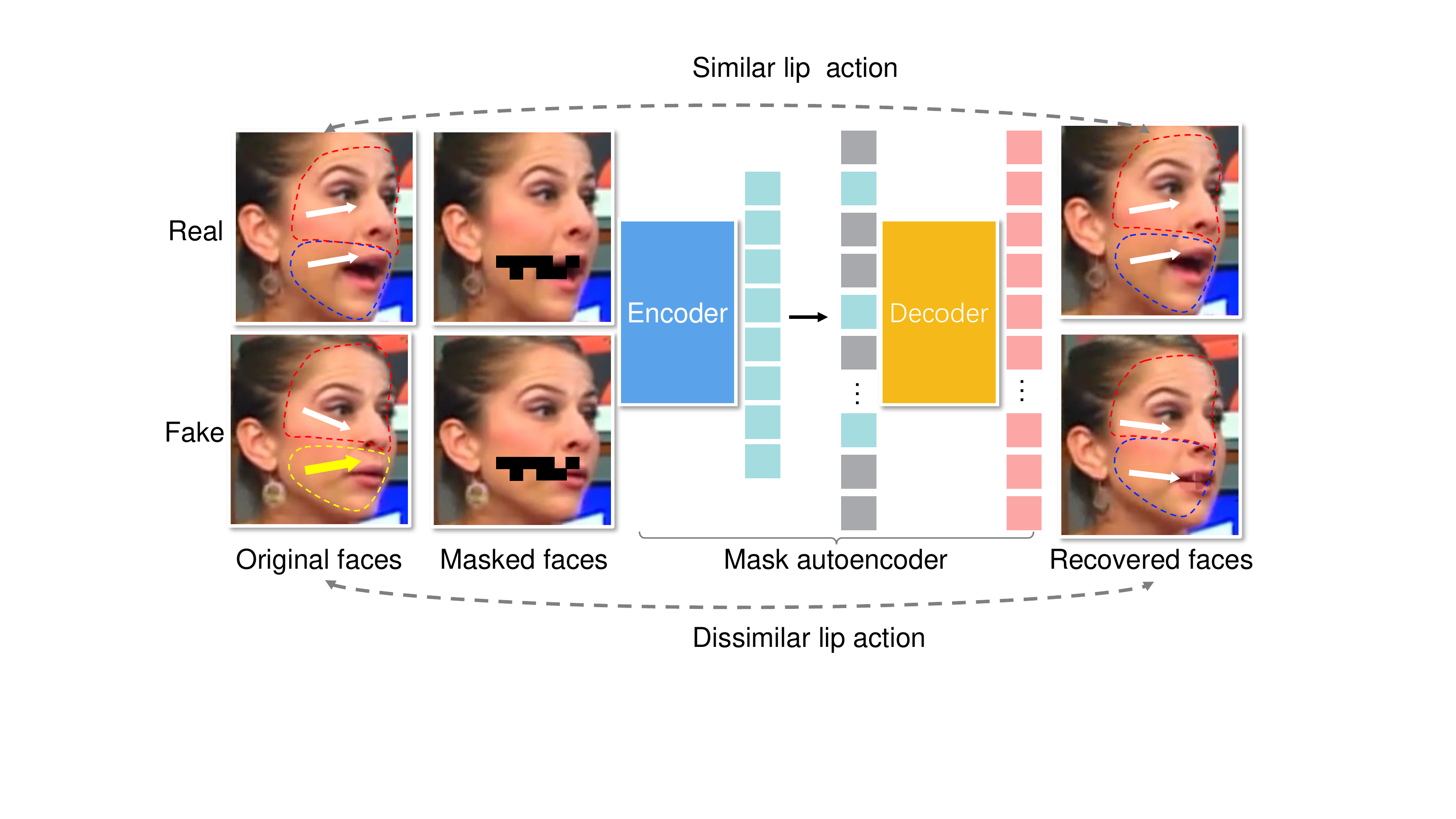}
\vspace{-0.4cm}
   \caption{ Deepfake videos often lack consistency among facial parts. Exploiting these inconsistencies, our \textsf{Mover} method effectively distinguishes real faces from fakes by recovering real faces more accurately than fake images, while being pretrained only on real images.}

   

   
   \label{fig1}
\vspace{-0.4cm}
\end{figure}

\section{Introduction}
Deepfake refers to the utilization of AI algorithms to fabricate convincing yet fake videos of a person's face -- e.g. speaking words that they have never actually done before \cite{chesney2019deep,aibase}. Despite its potential in fields like film and entertainment, malicious use of Deepfake technology poses significant threats to individuals and society \cite{fakefake}, emphasizing the importance of developing effective Deepfake detection techniques. 


Numerous techniques for detecting Deepfakes have been proposed, which can be broadly classified into two categories: those based on implicit clues and those based on explicit clues. Methods based on implicit clues~\cite{meso, capsule,tan2019efficientnet, ffdata, smil, ltw, CNN-GRU, multiatt, recon, transformer22mm, aaaidual, aaailocal} use supervised learning to distinguish genuine and fake videos without explicitly incorporating clues to detect Deepfake videos, making it challenging to understand the underlying detection clues. Recent methods that employ explicit clues~\cite{eye, head, thinking, frequency, cnn2, emotions, luo2021generalizing,on, hcil, xray, lips, spsl, finfer, stil, UIA-ViT, Patch-based, self, pcl, ftcn, delv, wavelet22mm,haliassos2022leveraging,dong2022protecting,shiohara2022detecting} have achieved more promising performance. However, given the rapid advancement of Deepfake technology, various falsification traces can be left behind, rendering detection methods that rely on specific facial features vulnerable to attack. Additionally, in real-world scenarios, the provided video may be generated using unknown manipulation techniques, which may result in a drop in performance~\cite{ffdata,tan2019efficientnet,xray,li2018learning,thinking,multiatt,luo2021generalizing,ltw,aaailocal,aaaidual} when detecting forgeries outside of the training dataset. This motivates us to develop a method that does not depend on manually designed clues or specific Deepfake patterns but also does not simply perform supervised learning to distinguish between genuine and fake.

Our idea, inspired by psychology and facial expression studies \cite{facs, mico1, micro2, expression, xinli}, is to harness the inherent consistency present in real human faces.  Since fake faces lack genuine human psychology, it becomes challenging to maintain consistency in facial parts. Consistency refers to the harmony and coherence among facial parts, such as eyes, eyebrows, and mouth, as they contribute to overall facial expressions. These consistencies exist in real faces but are often challenging to replicate through Deepfake generation methods \cite{head, stil, consis, hcil, UIA-ViT,yang2023masked}. For example, in Fig. \ref{fig1}, the fake face from the FaceForensics++ dataset \cite{ffdata} has been manipulated by Face2Face\cite{thies2016face2face} to  be upward, lacking consistency with other facial features because the cheek and eye parts are not upward. We can further confirm this inconsistency in fake videos by  checking the pixel-level correlation (Fig.\ref{fig4}) and the inconsistency areas of facial parts (Fig.\ref{fig5}).

How can we utilize facial consistency for robust Deepfake detection? We hypothesize that while it is difficult to generate videos with perfect facial consistency, creating a model that can restore a masked face image while maintaining consistency is easier and more feasible. To realize this restoration concept, we pretrain a simple yet powerful masked autoencoder (MAE) model~\cite{imgmae} solely on real images. Furthermore, we guide its masking pattern by facial parts (Algorithm~\ref{algorithm1}), enabling the model to learn unspecific facial part consistency. The consideration of the facial parts is important, which distinguishes our work from trying to reconstruct the whole face without considering the  semantic parts of faces~\cite{recon}. Consequently, applying our pretrained MAE to fake images with inconsistent facial parts yields more differences between the original and reconstructed fakes than real faces. In this manner, our MAE pretraining, which we refer \textsc{stage 1}, provides a powerful representation to distinguish real faces from Deepfake faces. We further finetune the pretrained model for more robust Deepfake detection by using Finetuning Network and Mapping Network to detect fakes for unknown domains, which we refer \textsc{stage 2}.

\textsc{Stage 1:} The primary objective  of the first stage is \textit{pretraining a model to capture the consistency among all facial parts in real faces}. Inspired by \cite{imgmae}, we propose a novel self-supervised pretraining method  that is based on \underline{m}asking and rec\underline{over}ing faces, named \textsf{Mover}. Unlike MAE~\cite{imgmae}, which randomly masks most areas of the image, \textsf{Mover} introduces a  masking strategy conditioned on facial part consistencies to randomly mask the Regions of interest (ROIs).  When we split the faces into different parts and randomly select a part to mask the ROIs randomly, the autoencoder can thus better learn the 
consistencies among all facial parts by reconstructing (or recovering) the missing pixels.

\textsc{Stage 2:} The goal of the second stage is \textit{utilizing the pretrained model to distinguish a real and fake video}. Specifically, it finetunes the representation from the first stage, so that it has better discriminability of real and fake ones and better domain generalization on diverse Deepfake patterns. To do so, we design two sub-networks. The first branch (Fig.\ref{fig2}-a) is  Finetuning Network that aggregates facial part consistencies and finetunes the model to classify real videos and fake videos. Using the trained encoder of the first stage to extract features, the model can learn the difference between real videos and fake videos with respect to the consistency of facial parts. The second branch (Fig.\ref{fig2}-b) is the Mapping Network, which employs the trained masked autoencoder of the first stage to reconstruct faces. By developing a Mapping Network, the branch can map original real faces into its \textsf{Mover}'s reconstruction but fail to map fake faces into its \textsf{Mover}'s reconstruction. By minimizing the mean squared error (MSE) loss between reconstructed faces and mapped faces, the model amplifies inconsistencies in fake faces and consistencies in real videos. Furthermore, to explicitly promote the generalization to unseen DeepFake patterns, we use meta-learning, where the different types of Deepfake videos are split into Meta-train and Meta-test in the training phase. The main contributions are as follows.

\noindent{(1)} We propose  \textsf{Mover} for the Deepfake detection that can learn to utilize the consistencies among all facial parts by masking and recovering faces, which presents an important clue to identifying unknown Deepfakes. To the best of our knowledge, we are the first to adapt  the masked autoencoder for Deepfake detection. We find that straightforwardly applying a masked autoencoder (MAE) for Deepfake detection did not achieve satisfactory results, but developing a unique adaptation of the MAE allows us to detect Deepfakes successfully. 

\noindent{(2)} To develop a model that can generalize to unseen Deepfake patterns, we introduce a two-stage strategy, where the first stage uses real images only and learns the representation unspecific to any Deepfake or any facial part, and the second stage learns a more robust representation that maximizes the discrepancy between real faces and fake faces. 

\noindent{(3)}  Extensive experiments on benchmark datasets, including FaceForensics++ \cite{ffdata}, Celeb-DF \cite{celeb}, WildDeepfake (WildDF) \cite{wild}, and DFDC preview (DFDCP) \cite{dfdcpre} show that \textsf{Mover} achieves excellent performance under various metrics.

\vspace{-0.2cm}

\section{Related Work}

\subsection{Deepfake Synthesis Methods}
Li et al. \cite{celeb} break down the Deepfake synthesis methods into two generations. First-generation methods include Face2Face \cite{thies2016face2face}, FaceSwap \cite{faceswapgit}, NeuralTextures \cite{ne}, and DeepFakes \cite{deepfakegit}. Face2Face synthesizes target faces by warping source faces. FaceSwap applies the rendered model and color correction to swap the face region. NeuralTextures trains a model to learn neural texture features of target videos and perform facial reenactment. DeepFakes is based on two autoencoders and a decoder to generate fake videos. First-generation methods have some known issues, such as color mismatches, temporal flickering, and incorrect face boundaries. The second-generation methods improve the quality over the previous methods. Specifically, Celeb-DF \cite{celeb} applied an algorithm to reduce boundary artifacts and improve inter-frame continuity. Dolhansky et al. \cite{dfdcpre} generate videos with low facial ratios using swap algorithms. 

\indent These diverse synthesis methods leave different tampered traces and cause biased data distributions, making it challenging to detect Deepfake videos in unknown domains. In this paper, we utilize the videos from public datasets that contain the aforementioned two generations' methods.

\vspace{-0.2cm}
\subsection{Deepfake Detection Methods}

Methods based on implicit clues mainly perform supervised learning to classify fake or real. The early method \cite{meso} develops Mesonet to extract mesoscopic properties. To amplify artifacts and suppress the high-level face content, Masi et al. \cite{twobranch} introduce a two-branch recurrent network for isolating manipulated faces. Zi et al. \cite{wild} propose ADDNets with attention mechanisms for advanced Deepfake detection. To improve the robustness, Cao et al. \cite{recon} is an early attempt to reconstruct the  face image, which captures  implicit clues and detects Deepfake videos by exploiting reconstruction differences.

\indent Methods based on explicit clues leverage various specific information to improve detection performance. Signal clues such as frequency information \cite{thinking, spsl}, local textures \cite{fwa}, and biological features \cite{fakecatcher} are employed to distinguish Deepfake videos from real videos. Semantic clues like emotion features between audio and visual \cite{emotions}, temporal inconsistencies \cite{consis, fttwostream}, and lips information \cite{lips}  exhibit remarkable progress in Deepfake detection. 
To  develop  generalizable Deepfake detectors, Zhao et al. \cite{pcl}, Zheng et al. \cite{ftcn}  and Guan et al. \cite{delv} propose reliable frameworks. Furthermore, Yang et al \cite{yang2023masked} extract features from facial regions and use the graph to model relationships of  facial regions. To address Deepfake detection, Shiohara et al. \cite{shiohara2022detecting} develop self-blended images (SBI) to detect Deepfakes, achieving the best generalization
ability when detecting Celeb-DF and DFDCP.

\indent These methods achieve promising results on current benchmarks, but they rely on specific clues. With the emergence of new Deepfake techniques, specific clues extracted by detection methods may be circumvented by purposely training during the synthesis of fake videos. Unknown Deepfake patterns leave complex manipulated traces rather than foreseeable specific features, which motivates us to further explore developing detectors for unknown domains  using unspecific clues. Hence, we propose a self-supervision method that reconstructs randomly masked faces to learn robust features that are not specific to any facial parts,  promoting the model to learn the consistencies of facial parts.
\vspace{-0.2cm}
\subsection{Masked Autoencoder Methods}
The emergence of the masked autoencoder (MAE)~\cite{imgmae} has greatly influenced our community. Thereafter, extent MAE \cite{vimae},  VideoMAE \cite{videomae}, and ConvMAE \cite{conmae} are proposed to learn strong representations. These MAE methods are based on an asymmetric encoder-decoder architecture. The extent MAE \cite{vimae} mainly extends MAE  to spatiotemporal representations. VideoMAE \cite{videomae} develops a customized design of tube masking based on MAE.  ConvMAE \cite{conmae} introduces a convolution structure in MAE. Although they are relatively small modifications, the performance of various downstream tasks has been greatly improved.

MAE \cite{imgmae} is designed for pretraining  Vision Transformers (ViT) rather than for Deepfake detection. MAE randomly masks patches with a high ratio, but our modified masking strategy differs from that of MAE in that it emphasizes facial part consistency. By doing so, the autoencoder is forced to learn a representation that maintains the consistency of facial parts.

\begin{figure*}[t]
  \centering
   \includegraphics[width=0.9\linewidth]{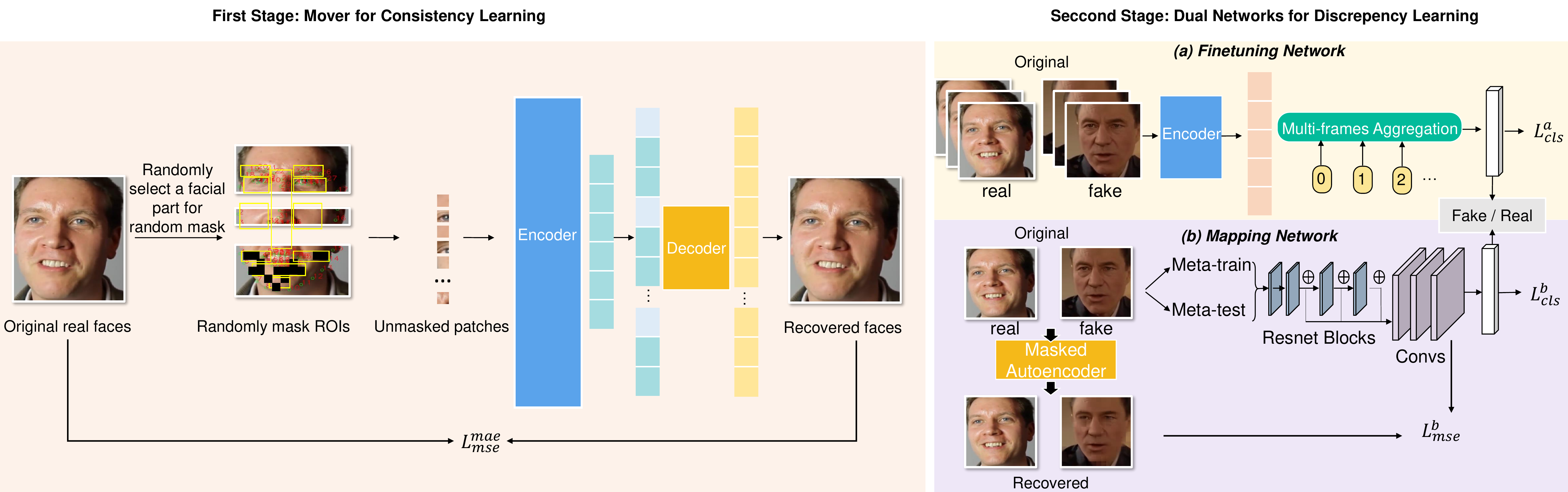}
\vspace{-0.4cm}
   \caption{Pipeline of the propose \textsf{Mover}. In the first stage, \textsf{Mover}  learns  general facial part consistencies of real faces by developing the masked autoencoder. In the second stage,  \textsf{Mover} performs \textit{(a) Finetuning Network}  and \textit{(b) Mapping Network}  to leverage the pretrained \textsf{Mover} model for better capturing the discrepancy between real videos and fake videos.}
   \label{fig2}
\vspace{-0.2cm}
\end{figure*}

\section{Method}

This section presents the details of \textsf{Mover} for Deepfake video detection. To better extract robust features for unknown Deepfake patterns, this method utilizes a facial masked autoencoder for feature learning. Furthermore,  the proposed method utilizes dual networks to detect differences between real videos and fake videos using the trained model of the facial masked autoencoder. Specifically, the proposed method is composed of two stages: (1) \textsf{Mover} for the Consistency Learning stage, (2) Dual Networks for the Discrepancy Learning stage, as shown in Fig.~\ref{fig2}. The first stage trains the model to learn the facial part consistencies of real faces. The second stage  is composed of two branches: Finetuning Network  and Mapping Network.

 {\renewcommand\baselinestretch{1.0}\selectfont

\begin{algorithm}[!t]

\caption{Masking process of facial parts}
\label{algorithm1}

\KwIn{Original faces: $F$,  Mask ratio: $M_r$;}

\KwOut{Masked  faces;}

Resize the face image with a size of 224*224, and detect $68$ keypoint landmarks in faces;

Define $ROIs=\{R_1,R_2...R_{11}\}$.

Define $196$ blocks of face image;

$M =\{M_1, M_2, M_3\} $, $M_1:$ eyes patches $\in$ \{$(0,0)$  $\rightarrow$  $17^{th}_{landmark}$\}, and $M_2:$ Nose \& cheek patches $\in$ \{end of eyes patches   $\rightarrow$  $16^{th}_{landmark}$\}, and $M_3:$ lips patches $\in$ \{end of Nose \& cheek patches   $\rightarrow$ the last patches of $F$\};

 \For {$i = 1$ \textbf{to} $3$}
    {Randomly select $M_i$
    
     \For {$block \in selected$ \indent $M_{i}$}
     {\If {$ROI$ \indent$landmarks \in block$}
    {Sign the blocks. }
    { Number of mask patches  = signed blocks $\times$ $M_r$;
     
    Random mask the patches. }}}

\end{algorithm}

\par}

\subsection{\textsf{Mover} for Consistency Learning}\label{sec:mafpc}

In this stage, we customize MAE\cite{imgmae} and its masking strategy with  modifications to perform self-supervised learning of real faces, so that the model can learn generic facial part consistency features and can also prevent over-fitting to specific forgery patterns.

\noindent \textbf{Masking strategy tailored to learn the consistent face representation. }We customize the original masking strategy from MAE~\cite{imgmae} and design a facial part masking strategy to ensure that the model can learn the consistencies of all facial parts. We show the pseudo-code in Algorithm \ref{algorithm1}, which takes the inputs of videos' faces  and outputs the masked faces. 
The original MAE~\cite{imgmae}  masks random patches of the input image, but the same strategy is not suitable for our \textsf{Mover} for the following reasons. 

\indent First, the original MAE masking strategy, with random masking without considering  Deepfake's domain knowledge, can not be directly utilized for Deepfake detection. This is because randomly masking pixels causes the model to focus solely on local facial consistency, ignoring global consistency across different facial parts. Such a lack of global facial part consistency can impede the model's ability to learn accurate facial part consistency features, leading to difficulties in detecting differences between real and fake videos from reconstructed faces (Fig. \ref{fig5}). 
Therefore, our proposed masking strategy explicitly splits faces into three facial parts, i.e., eyes, cheek \& nose, and lips, enabling the model to focus on both local and global consistencies among all facial parts.  We note that the ROIs  extraction  is partially inspired  by Facial Action Coding System (FACS) \cite{facs}, which considers the action units of FACS as fundamental elements of facial expressions. Based on psychology and facial expression studies \cite{facs, mico1, micro2, expression, xinli}, the real faces exhibit  inherent consistency in these elements. Consequently, masking ROIs enables more consistent reconstruction of these regions for real faces compared to fake faces. 
We reference the action units of eyebrows,  lower eyelid, nose root, cheeks, mouth corner, side of the chin, and  chin to calculate bounding box coordinates according to facial keypoints. Specifically,  $R_1$ and $R_2$ are the eyebrows, and $R_3$ and $R_4$ are the lower eyelid, and $R_5$ is the nose root, and $R_6$ and $R_7$ are the cheeks, and $R_8$ and $R_9$ are the mouth corner, and $R_{10}$ is the side of the chin, and $R_{11}$ is the chin.

\indent Second, the original MAE masking strategy, with a high masking ratio, would make it too challenging to restore the original appearance without any artifacts or distortions. If reconstruction artifacts occur, real faces will contain them, and fake faces will have both reconstructed artifacts and tampering artifacts. This makes it difficult to distinguish between real and fake videos since both have artifacts. Therefore, we propose a masking strategy that utilizes a relatively low masking ratio to enable the model to reconstruct the original faces more accurately. We discuss more about the masking strategy in Section \ref{ablation}.

 \noindent \textbf{Network Architecture.}  Our masked autoencoder is based on an asymmetric encoder-decoder architecture \cite{imgmae}. The encoder is a ViT \cite{vit} but applied only on unmasked patches. The decoder is a series of Transformer blocks applied on both mask tokens and encoded features of unmasked patches.
 \\
   \textbf{Recover masked faces.} The masked patches of faces are dropped in the processing of the encoder, leaving the unmasked areas. In this way, the decoder predicts the missing facial part based on the unmasked areas. The reconstruction quality of masked patches is calculated with the MSE loss function $L^{mae}_{mse}$, which can be represented as: 
   \begin{equation}
L^{mae}_{mse} = \frac{1}{n}\sum^{n}_{j=1}{{(\widetilde{y^j}-y^j)}^2},
 \end{equation}
where, $y^j$ is the ground truth pixel, $\widetilde{y^j}$ is the predicted pixel, and $n$ represents the number of the face pixel. If the model learns consistencies among facial parts, the loss between the reconstructed patches and the input patches should decrease. Our facial part masking strategy makes each part selected randomly, which enforces the model to learn the representation unspecific to any facial part. Furthermore, because this stage only uses real videos and does not use any Deepfake videos, it can prevent the model from over-fitting to any specific tampering pattern. After the training of the first stage, the pretrained \textsf{Mover} model is obtained.

\vspace{-0.3cm}
\subsection{Dual Networks for Discrepancy Learning}\label{sec:stage2}
In this stage, we leverage the well-trained \textsf{Mover} model from the first stage so that the model can better capture the discrepancy between real and fake videos, using two branches: (a) Finetuning Network, and (b) Mapping Network. By adopting  the well-trained model of the first stage, the two branches are combined to detect Deepfake videos with separate training.

\indent (a) Finetuning Network. Finetuning Network uses both real videos' frames and  Deepfake videos' frames with a cross-entropy loss $L_{cls}^{a}$ to extract the discrepancy between real videos and fake videos. For $N$ samples, $L_{cls}^{a}$ can be calculated as:
   \begin{equation}
L_{cls}^{a} = \frac{1}{N}\sum_{m}{-[g_m^a log(p_m^a)+(1-g_m^a)log(1-p_m^a) ]},
 \end{equation}
where $g_m^a$ is the ground truth, and $p_m^a$ is the probability that the prediction is the label of real video.

We first extract frames from real and Deepfake videos, and crop faces from frames. Thereafter, original real faces and original fake faces with full sets of patches are put into the well-trained encoder of the first stage. To aggregate the information of multi-frames, we average the outputs of $10$ frames, using the last layer of the encoder. 

\indent Since the first stage learns the facial part consistency of real videos, the well-trained encoder of \textsf{Mover} can extract consistency features of real videos. For fake videos, because the consistency is destroyed, the features extracted from the encoder can be different from that of real videos. Finetuning Network will amplify the discrepancies between fake and real videos, which improves the detection performance.

\indent(b) Mapping Network. Mapping Network contains the following components.

\noindent \textbf{Data split strategy.} To avoid over-fitting to specific Deepfake patterns, we use meta-learning \cite{meta} and randomly divide the training data into Meta-train set and Meta-test set, where fake faces in Meta-train and Meta-test have different manipulated patterns. Since this branch requires different types of faces rather than a large number of faces, we utilize a single frame to train the branch for reducing  memory consumption.\\
   \textbf{Network Architecture. } We utilize the first convolutional layer of ResNet-18 \cite{resnet}.  The first three residual blocks of ResNet-18 are employed, and the outputs of these residual blocks are concatenated. The concatenated outputs are fed into three convolutional layers for face mapping. The dimensions of the mapped faces are $56*56*3$. We minimize the MSE loss $L_{mse}^{b}$ between the mapped faces and the reconstructed faces of the input. Ultimately, a fully connected (FC) layer is utilized for classifications. We also minimize the binary cross-entropy $L_{cls}^{b}$ between the branch output and the video label.
\\ \textbf{Meta-train phase.} For each epoch, a sample batch is formed with the same number of fake videos and real videos to construct the binary detection task. The Meta-train phase performs training by sampling many detection tasks, and is validated by sampling many similar detection tasks from the Meta-test. Thereafter, the parameters of Meta-train phase can be updated. 
To select the best gradient step, we set a reference set, denoted as $T_i^{ref}$. We utilize each gradient step to calculate the accuracy of $T_i^{ref}$. The parameters with the highest accuracy are selected as the final updated parameters. Ultimately, the Meta-train phase utilizes the updated parameters to calculate the Meta-train loss.

\indent Since the first stage learns the consistencies of all facial parts in real videos, the model should output inconsistent representations when the consistency of a face is destroyed. In other words, the fake faces reconstructed from the trained masked autoencoder of the first stage should expose inconsistencies. Optimizing $L_{mse}^{b}$ allows the mapped faces of real videos to be constrained by the consistency of reconstructed real faces, while the mapped faces of fake videos are constrained by the inconsistency of reconstructed fake faces.  Ultimately, the model can observe that the real faces maintain consistencies in facial parts, while the fake ones'  consistencies are broken.
\\\textbf{Meta-test phase. }The goal of Meta-test phase is to enforce a classifier that performs well on Meta-train and can quickly generalize to the unseen domains of Meta-test, so as to improve the cross-domain detection performance. Specifically, we sample a batch in Meta-test domain and another batch in the Meta-train domain to concatenate to a random array. Then, we use the random array and the updated parameters to compute the Meta-test loss of $L_{cls}^{b}$ and $L_{mse}^{b}$.
\\\textbf{Detection loss. }The final loss function of Mapping Network is:
\begin{equation}
L^b =  (  L^b_{cls} + L^b_{mse} )_{Meta-train}+( L^b_{cls} + L^b_{mse} )_{Meta-test}.
\end{equation}
which combines the Meta-test loss of $L_{cls}^{b}$ and $L_{mse}^{b}$ and Meta-train loss of $L_{cls}^{b}$ and $L_{mse}^{b}$ to achieve joint optimization.


\section{Experiment}

\subsection{Experimental Setup}
\indent \textbf{Datasets.} Four public Deepfake videos datasets, i.e.,  FaceForensics++ \cite{ffdata}, Celeb-DF  \cite{celeb}, WildDF \cite{wild}, DFDCP \cite{dfdcpre} are utilized to evaluate the proposed method and existing methods. FaceForensics++ is made up of $4$ types manipulated algorithms: DeepFakes \cite{deepfakegit},  Face2Face \cite{thies2016face2face}, FaceSwap \cite{faceswapgit}, NeuralTextures \cite{ne}. Moreover, $4000$ videos are synthesized based on the $4$ algorithms. These videos are widely used in various Deepfake detection scenarios. Celeb-DF contains $5639$ videos that are generated by an improved DeepFakes algorithm \cite{celeb}. The tampered traces in some inchoate datasets are relieved in Celeb-DF. WildDF consists of $707$ Deepfake videos that were collected from the real world. The real world videos contain diverse scenes, facial expressions, and forgery types, which makes the datasets challenging. DFDCP   is a large-scale Deepfake detection dataset published by Facebook.

\noindent \textbf{Implementation details.} In the first stage, the masking ratio, batch size, patch size, and input size are set as $0.5$, $8$, $16$, $224$, respectively. The AdamW \cite{adamw} optimizer with an initial learning rate $1.5 \cdot 10^{-4}$, momentum
of $0.9$ and a weight decay $0.05$ is utilized to train the model. In the second stage, the Finetuning Network utilizes the AdamW  optimizer with an initial learning rate $1 \cdot 10^{-3}$ to detect videos. The
SGD optimizer is used for optimizing the Mapping Network with the initial learning rate $0.1$, momentum of $0.9$, and weight decay of $5\cdot 10^{-4}$. We use \texttt{FFmpege} \cite{ffmpeg} to extract frames from videos. The \texttt{dlib} \cite{dlib} is utilized to detect 68 facial landmarks. 
We randomly mask facial parts according to Algorithm \ref{algorithm1} for the training of the first stage. In the second stage, we use \texttt{dlib} to extract faces. These faces are fed into the model without masking patches. 
\begin{figure}
 \centering
\includegraphics[width=0.8\linewidth]{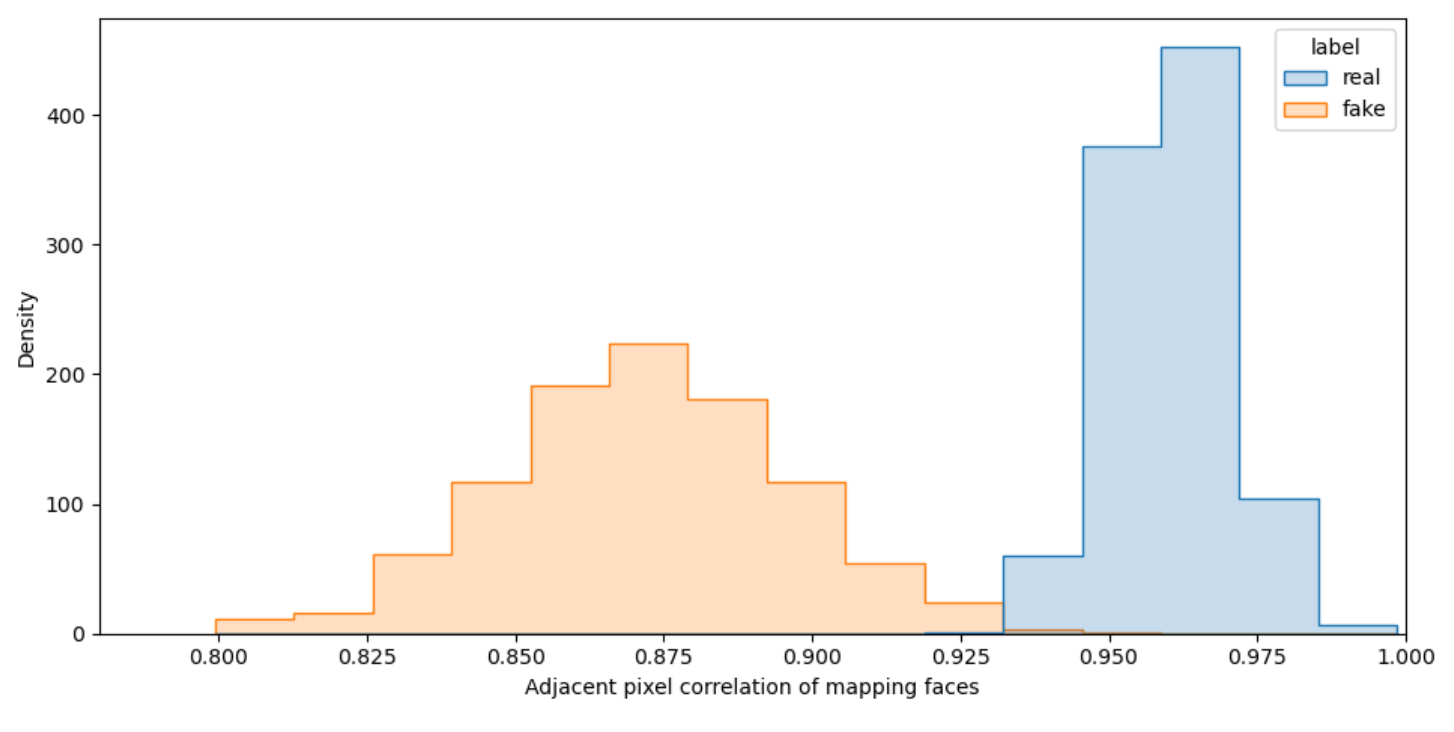}
  \vspace{-0.2cm}
\caption{The distinction in  the pixel-level correlation of the mapped faces.}
 \label{fig4}
  \vspace{-0.5cm}
\end{figure}
\vspace{-0.2cm}
\subsection{Preliminary Analyses of the Inconsistency}

To better understand how \textsf{Mover} effectively detects inconsistencies in fake faces, we conduct a simple experiment utilizing its trained Mapping Network (Sec~\ref{sec:stage2}(b)). Recall that the Mapping Network is designed to map recovered faces into faces that emphasize the differences between recovered real faces and recovered fake faces. Thus, by examining the mapped faces of real and fake samples, we may be able to quantify the inconsistencies. We empirically discover that the simple Pearson correlation scores between each pixel and its neighboring pixels can demonstrate these inconsistencies.

In Fig. 3, we present statistics from 1000 randomly sampled real and fake videos, derived from datasets featuring both whole-face and partial-face swaps, such as FaceForensics++, Celec-DF, WildDF, and DFDCP. The results reveal that the adjacent pixel correlation scores of mapped real faces consistently exceed $0.91$, indicating that real faces maintain facial consistency throughout the reconstruction process. This can be attributed to \textsf{Mover}'s first stage, which accurately reconstructs real faces, while the $L_{mse}^{b}$ constraint in the second stage aids in preserving facial consistency, resulting in high correlation scores. In contrast, the adjacent pixel correlation scores of mapped fake faces are considerably lower due to inconsistencies in the reconstructed faces. As a consequence, the $L_{mse}^{b}$ constraint prompts the mapped face to align with the inconsistent features of the fake face, leading to low correlation scores. The inconsistency is present in both whole-face and partial-face swaps. Consequently, when we recover faces based on unmasked areas, any inconsistency in any part can contribute to an overall inconsistent reconstruction. We also assess the performance on multiple datasets involving full face swap and partial face swap scenarios, which we discuss in the following section.

\vspace{-0.2cm}
\subsection{Generalization to Unknown Domains}
 We design a detector that automatically learns consistencies among all facial parts, which enforces the model to learn robust features for the detection of unknown domains. To test this point, we simulate unknown domain Deepfake detection scenarios below.

\begin{table}[!t]
\begin{center}
\vspace{-0.4cm}
\caption{{Cross-dataset generalization. }Comparisons of the cross-dataset evaluation  (EER and AUC (\%)) between \textsf{Mover} and baseline methods on Celeb-DF, WildDF, and DFDCP datasets when trained on FaceForensics++. }
\vspace{-0.3cm}
\label{crossdatasets}
\renewcommand{\tabcolsep}{0.05mm} 
\renewcommand{\arraystretch}{1.0}
\begin{tabular}{ccccccccc}
\noalign{\hrule height 0.7pt}


     \multirow{2}{*}{Method}
     &\multicolumn{2}{c} {Celeb-DF}&\multicolumn{2}{c} {WildDF}& \multicolumn{2}{c} {DFDCP}&\multicolumn{2}{c} {Avg}\\ 
    \cmidrule(r){2-3} \cmidrule(r){4-5}\cmidrule(r){6-7}\cmidrule(r){8-9}&AUC $\uparrow$ &EER$\downarrow$ &AUC $\uparrow$ &EER$\downarrow$&AUC $\uparrow$ &EER$\downarrow$&AUC $\uparrow$ &EER$\downarrow$\\
     
              \noalign{\hrule height 0.7pt}


{\textmd{Xception \cite{ffdata}}}&$65.3$&$38.8$&$66.2$&$40.1$& $69.8$&$35.4$&$67.1$&$38.1$ \\
{\textmd{EN-b4 \cite{tan2019efficientnet}}}&$68.5$&$35.6$&$61.0$&$45.3$& $70.1$&$34.5$&$66.5$&$38.5$ \\
{\textmd{Face X-ray \cite{xray}}}&$74.2$&$-$&$-$&$-$& $70.0$&$-$&$-$&$-$ \\
{\textmd{MLDG \cite{li2018learning}}}&$74.6$&$30.8$&$64.1$&$43.3$& $71.9$&$34.4$&$70.2$&$36.2$ \\
{\textmd{$F^3$-Net \cite{thinking}}}&${71.2}$&$34.0$&${67.7}$&$40.2$& ${72.9}$&$33.4$&${70.6}$&$35.9$ \\

{\textmd{MultiAtt \cite{multiatt}}}&$76.7$&$32.8$&$70.2$&$36.5$& $67.3$&$38.3$&$71.4$&$35.9$ \\
{\textmd{GFF \cite{luo2021generalizing}}}&$75.3$&$32.5$&$73.0$&$34.9$& $71.8$&$31.8$&$73.4$&$33.1$ \\
{\textmd{LTW \cite{ltw}}}&$77.1$&$29.3$&$67.1$&$39.2$& $74.6$&$33.8$&$72.9$&$34.1$ \\
{\textmd{Local-relation \cite{aaailocal}}}& ${78.3}$&$29.7$&$68.8$&$37.5$& $76.5$&$32.4$&$74.5$&$33.2$\\
{\textmd{DCL \cite{aaaidual}}}& ${82.3}$&$26.5$&$71.1$&$36.2$& $76.7$&$32.0$&$76.7$&$31.6$\\

{\textmd{Li et al. \cite{wavelet22mm}}}& ${84.8}$&$21.9$&$73.8$&$31.8$& $78.5$&$29.1$&$79.0$&$27.6$\\






 
 {\textmd{EB4+SBI \cite{shiohara2022detecting}}}& $\textbf{90.1}$& ${18.2}$&$71.2$&$34.2$& $\textbf{85.1}$&$\textbf{23.2}$&$82.1$&$25.2$\\
 {\textmd{R34+SBI \cite{shiohara2022detecting}}}& ${86.1}$& ${20.8}$&$67.4$&$36 .7$& ${81.2}$&$26.1$&$78.2$&$27.9$\\
 \rowcolor{gray!20}
{\textsf{Mover}}
&${{87.1}}$&$\textbf{{17.2}}$&${\textbf{82.0}}$&${\textbf{25.9}}$&${{78.9}}$&$27.1$&${\textbf{82.7}}$&$\textbf{23.4}$\\

\noalign{\hrule height 0.7pt}

\end{tabular}
\end{center}
\vspace{-0.6cm}
\end{table}

\noindent \textbf{Cross-dataset generalization. }To  evaluate the generalization of \textsf{Mover}, we  construct cross-dataset experiments. Since videos of different datasets possess  relatively  dispersed distributions, cross-dataset experiments can simulate unknown domain detection. We follow the work of  Li et al. \cite{wavelet22mm} and implement the cross-dataset experiments by training the model on FaceForensics++ with all $4$ types of videos, but testing on other datasets, i.e., Celeb-DF, WildDF, DFDCP.

\indent Some results of Table \ref{crossdatasets} are cited from Li et al. \cite{wavelet22mm}, and we use Receiver Operating Characteristic Curve (AUC) and Equal Error Rate (EER) to evaluate the performance following Li et al. \cite{wavelet22mm}. EB4$+$SBI \cite{shiohara2022detecting} achieves the state-of-the-art cross-dataset performance in their paper, and we re-implement  EB4$+$SBI and report the results in Table \ref{crossdatasets}. The  detection performance of SBI \cite{shiohara2022detecting}  is based on a pre-trained  EcientNet-b4 model. We also re-implement R34$+$SBI proposed in Ref. \cite{shiohara2022detecting}.  When testing Celeb-DF and DFDCP, the AUCs of EB4$+$SBI \cite{shiohara2022detecting} are  higher than ours. However,   the  AUC and EER of  EB4$+$SBI \cite{shiohara2022detecting} are  lower than ours when testing WildDF. The main reason is that Celeb-DF focuses on one technology and DFDCP focuses on two technologies, while videos in  WildDF are collected from the real world and generated by various technologies. \textsf{Mover}'s ability to learn general representations that fit various videos makes it slightly distant from a single technology in order to fit multiple technologies. The enormous differences between the training domain and the testing domain make it challenging to improve cross-dataset detection performance. Nonetheless, the proposed \textsf{Mover} manages to reach slight advantages in terms of average AUC and EER.




\begin{table}
\setlength{\abovecaptionskip}{0.1cm}
\setlength{\belowcaptionskip}{0.1cm}
\tabcolsep=2pt
\begin{center}
\vspace{-0.5cm}
\caption{Cross-manipulation generalization. Comparisons of the cross-manipulation evaluation  (AUC (\%)) between \textsf{Mover} and baseline methods on each forgery type of FaceForensics++ when trained on one type. DeepFakes, FaceSwap, Face2Face, NeuralTextures are represented as DF,  FS,  F2F, and  NT, respectively.}
\label{crossffone}
\renewcommand{\tabcolsep}{2mm} 
\renewcommand{\arraystretch}{1.0}
\begin{tabular}{cccccc}
\noalign{\hrule height 0.7pt}
     {Method}& {Train}&{F2F}& {FS}&{NT}&{Avg}\\
           
\noalign{\hrule height 0.7pt}

{\textmd{Freq-SCL \cite{frequency}}}&$ \multirow{6}{*}{DF}$ & $58.9$&$66.9$ &$63.6$&$63.1$ \\
{\textmd{MultiAtt \cite{multiatt}}}&$ $&$66.4$& $67.3$ & $66.0$&$66.6$\\
{\textmd{RECCE \cite{recon}}}&$ $&${70.7}$& $74.3$ & $67.3$&$70.8$\\
{\textmd{LipForensics \cite{lips}}}&$ $&${70.0}$& $41.9$ & $76.1$&$62.7$\\

{\textmd{FInfer \cite{finfer}}}&$ $&${69.4}$& ${74.2}$ & ${66.8}$&${70.1}$\\
{\textmd{Freq-SCL \cite{frequency}}}&$ $ & $58.9$&$66.9$ &$63.6$&$63.1$ \\
 \rowcolor{gray!20}
{\textsf{Mover }}&$ $&${\textbf{86.9}}$& ${\textbf{77.6}}$ & ${\textbf{69.7}}$&${\textbf{78.1}}$\\
 \hline
     {Method}& {Train}&{DF}& {FS}&{NT}&{Avg}\\
           \hline

{\textmd{Freq-SCL \cite{frequency}}}&$ \multirow{6}{*}{F2F}$ & $67.6$&$55.4$ &$66.7$&$63.2$ \\
{\textmd{MultiAtt \cite{multiatt}}}&$ $&$73.0$& $65.1$ & $71.9$&$70.0$\\
{\textmd{RECCE \cite{recon}}}&$ $&${76.0}$& $64.5$ & ${\textbf{72.3}}$&$71.0$\\
{\textmd{LipForensics \cite{lips}}}&$ $&${88.4}$& $64.4$ & $70.2$&$74.3$\\

{\textmd{FInfer \cite{finfer}}}&$ $&${{76.3}}$& ${64.4}$ & $69.8$&${70.2}$\\

 \rowcolor{gray!20}
{\textsf{Mover }}&$ $&${\textbf{89.1}}$& ${\textbf{72.0}}$ & $65.5$&${\textbf{75.5}}$\\
 \hline
     {Method}& {Train}& {DF}& {F2F}&{NT}&{Avg}\\
           \hline
{\textmd{Freq-SCL \cite{frequency}}}&$\multirow{6}{*}{FS} $ & $75.9$&$54.6$ &$49.7$&$60.1$ \\

{\textmd{MultiAtt \cite{multiatt}}}&$ $&$82.3$& $61.7$ & $54.8$&$66.3$\\
{\textmd{RECCE \cite{recon}}}&$ $&${{82.4}}$& $64.4$ & ${56.7}$&$67.8$\\
{\textmd{LipForensics \cite{lips}}}&$ $&${60.9}$& $62.6$ & $57.9$&$60.5$\\

{\textmd{FInfer \cite{finfer}}}&$ $&${82.0}$& ${62.4}$ & $55.6$&$66.7$\\

 \rowcolor{gray!20}
{\textsf{Mover }}&$ $&${\textbf{82.6}}$& ${\textbf{82.9}}$ & ${\textbf{58.3}}$&${\textbf{74.6}}$\\
 \hline
     {Method}& {Train}& {DF}& {F2F}&{FS}&{Avg}\\
           \hline

{\textmd{Freq-SCL \cite{frequency}}}&$\multirow{6}{*}{NT} $& $79.1$&$74.2$ &$54.0$&$69.1$ \\
{\textmd{MultiAtt \cite{multiatt}}}&$ $&$74.6$& $80.6$ & $60.9$&$72.0$\\
{\textmd{RECCE \cite{recon}}}&$ $&${78.8}$& ${\textbf{80.9}}$ & ${63.7}$&${74.5}$\\
{\textmd{LipForensics \cite{lips}}}&$ $&${\textbf{90.0}}$& $51.9$ & ${\textbf{81.7}}$&$74.5$\\

{\textmd{FInfer \cite{finfer}}}&$ $&${{79.6}}$& $75.8$ & ${64.7}$&${73.4}$\\

 \rowcolor{gray!20}
{\textsf{Mover }}&$ $&${{86.8}}$& $69.4$ & $67.9$&${\textbf{74.7}}$\\
\noalign{\hrule height 0.7pt}
\end{tabular}
\end{center}
\vspace{-0.6cm}
\end{table}

\noindent \textbf{Cross-manipulation generalization. }We also carry out cross-manipulation experiments to  assess the generalization to unknown manipulation patterns. 
Videos in FaceForensics++ are generated from $4$ forgery technologies. Following RECCE \cite{recon}, we select one type of video for training and the remaining three types of videos for testing. 

\indent Results in Table \ref{crossffone} illustrate that \textsf{Mover} outperforms previous methods in many scenarios on average. Specifically, when the model is trained on DeepFakes and tested on the other three types of videos, \textsf{Mover} improves the performance by an average of 7.3\%.   When the model is trained on FaceSwap and tested on the other three types of videos, \textsf{Mover} improves the performance by 6.8\% on average. However, it should be noted that we re-implement LipForensics \cite{lips} and find that LipForensics \cite{lips} performs better than ours when training on NeuralTextures and testing on DeepFakes and FaceSwap, and RECCE \cite{recon} outperforms \textsf{Mover} when training on NeuralTextures and testing on Face2Face, as well as when training on Face2Face and  testing on NeuralTextures.  Despite these results, it is worth noting that the proposed \textsf{Mover} achieves comparable performance on average.

\subsection{Intra-dataset Detection Performance} To provide a comprehensive assessment of the proposed \textsf{Mover}, we compare \textsf{Mover} with the state-of-the-art methods in the scenario of intra-dataset detection. Specifically, we conduct experiments on $4$ subsets of FaceForensics++ (C23). The training data and testing data of intra-dataset experiments are  from the same subset of FaceForensics++. Table \ref{intra} shows  that most methods perform well in intra-dataset detection.  \textsf{Mover} achieves the highest intra-dataset detection score on Face2Face while having a slight decrease of  $0.9\%$ in average accuracy compared to LipForensics\cite{lips}, which ranks highest on average among the evaluated approaches. In terms of cross-manipulation detection performance, as shown in Table \ref{crossffone}, \textsf{Mover} surpasses LipForensics\cite{lips} in multiple metrics.

\begin{table}
\setlength{\abovecaptionskip}{0.1cm}
\setlength{\belowcaptionskip}{0.1cm}
\tabcolsep=2pt
\vspace{-0.4cm}
\begin{center}
\caption{Comparisons of the Intra-dataset evaluation  (AUC (\%)) between \textsf{Mover} and baseline methods on FaceForensics++.}
\label{intra}
\renewcommand{\tabcolsep}{2mm} 
\renewcommand{\arraystretch}{1.0}
\begin{tabular}{cccccc}
\noalign{\hrule height 0.7pt}
     {Method}& {DF}&{FS}& {F2F}&{NT}&{Avg}\\
           
\noalign{\hrule height 0.7pt}

 
 
 
 

 {\textmd{Freq-SCL \cite{frequency}}}&${\textbf{100}}$&${\textbf{100}}$& ${99.3}$& ${98.0}$&${99.3}$ \\
{\textmd{MultiAtt \cite{multiatt}}}&$99.6$&${\textbf{100}}$& ${99.3}$& ${98.3}$&$99.3$\\
{\textmd{RECCE \cite{recon}}}&$99.7$&${{99.9}}$& $99.2$ & ${98.4}$&$99.3$\\
{\textmd{LipForensics \cite{lips}}}&${\textbf{99.8}}$&${\textbf{100}}$& $99.3$ & ${\textbf{99.7}}$&${\textbf{99.7}}$\\

{\textmd{FInfer \cite{finfer}}}&$98.4$&${96.0}$& ${93.5}$ & $94.9$&$95.7$\\

  \rowcolor{gray!20}
{\textsf{Mover}}
&${99.6}$&$99.7$&${\textbf{99.6}}$&$96.3$&$98.8$\\

\noalign{\hrule height 0.7pt}
\end{tabular}
\end{center}
\vspace{-0.6cm}
\end{table}

\begin{table}
\setlength{\abovecaptionskip}{0.1cm}
\setlength{\belowcaptionskip}{0.1cm}
\tabcolsep=2pt
\begin{center}
\caption{ AUC (\%) scores of the FaceForensics++ testset  that are subject to different processing operations.}
\label{operation}
\renewcommand{\tabcolsep}{1mm} 
\renewcommand{\arraystretch}{1.0}
\begin{tabular}{ccccc}
\noalign{\hrule height 0.7pt}
     {Method}& {Resizing}& {Enhancement}&{Brightness}&{Contrast}\\
           
\noalign{\hrule height 0.7pt}
 {\textmd{R34+SBI \cite{shiohara2022detecting}}}& ${93.5}$& ${87.5}$&$83.2$&$84.5$ \\
 {\textmd{EB4+SBI \cite{shiohara2022detecting}}}& ${97.1}$& ${90.8}$&$87.6$&$87.4$ \\

  \rowcolor{gray!20}
{\textsf{Mover}}&${\textbf{97.8}}$&${\textbf{91.0}}$ &${\textbf{89.9}}$ &${\textbf{90.3}}$ \\
\noalign{\hrule height 0.7pt}

\end{tabular}
\end{center}
\vspace{-0.5cm}
\end{table}

\begin{table}
\setlength{\abovecaptionskip}{0.1cm}
\setlength{\belowcaptionskip}{0.1cm}
\tabcolsep=2pt
\begin{center}
\caption{Ablation study - The cross-dataset detection performance (AUC (\%)) of different masking ratios (55\%, 65\%,75\%, 85\%, 95\%) on Celeb-DF, WildDF, and DFDCP datasets.}
\label{ratio}
\renewcommand{\tabcolsep}{4mm} 
\renewcommand{\arraystretch}{1.0}
\begin{tabular}{cccc}
\noalign{\hrule height 0.7pt}
     {Mask ratio}& {Celeb-DF}& {WildDF}&{DFDCP}\\
           
\noalign{\hrule height 0.7pt}
\textmd{55\%}&$85.4$&${80.3}$ &$76.6$  \\
 
\textmd{65\%}&$86.5$&${81.6}$ &$77.3$  \\
  \rowcolor{gray!20}
\textmd{75\%}&${\textbf{87.1}}$&${\textbf{82.0}}$ &${\textbf{78.9}}$  \\
 
\textmd{85\%}&$86.1$&${81.7}$ &$77.6$  \\
 
 \textmd{95\%}&$85.8$&${80.1}$ &$76.9$  \\
\noalign{\hrule height 0.7pt}

\end{tabular}
\end{center}
\vspace{-0.7cm}
\end{table}

\subsection{Robustness to  Post-Processing Operations}
In the real-world situation, the frames and videos are often post-processed to adjust the media content for better display. The operations such as image resizing, image enhancement, video brightness, and video contrast are widely used. We post-precessing the frames or videos by using these operations and show the performance in Table \ref{operation}. After post-processing, specific forgery clues are relieved, thus SBI\cite{shiohara2022detecting} demonstrates sub-optimal results. While our method considers the unspecific facial part consistency, thus maintaining better detection performance even after post-processing.

\subsection{Ablation Study}
\label{ablation}

\indent \textbf{Influence of the masking ratio.} To evaluate the impact of the masking ratio on the generalization ability, we conducted experiments on the Celeb, WildDF, and DFDCP datasets, treating them as unseen datasets. We trained models on the FaceForensics++ dataset with different masking ratios.

\indent Note that instead of defining the masking ratio as the ratio of masked area to the entire face, we define the masking ratio as the ratio of masked area to the corresponding ROIs facial parts, as illustrated in Algorithm \ref{algorithm1}.  The reason why we do not use the original definition of mask ratio, that is, the ratio of the mask area to the whole face, is that we focus on ROIs and split faces into three parts and only randomly mask one part of the whole face at one time. We are supposed to focus on the corresponding masked ROIs parts in the masking process rather than the whole face. 

\indent In Table \ref{ratio}, we observe that \textsf{Mover} scales well with the masking ratio of 75\%.  The performance gets a slight drop in the masking ratio of 55\% and 65\% indicating that low masking ratios may hinder learning robust features. When the mask rate is 85\% and 95\%, the detection performance is also degraded. That may be because that high  masking ratio can raise the difficulty of reconstructing faces. If both real faces and fake faces are not reconstructed well, the distinction between them can be reduced. Therefore, we set the masking ratio as 75\%  in the experiments.

\begin{table}
\setlength{\abovecaptionskip}{0.1cm}
\setlength{\belowcaptionskip}{0.1cm}
\tabcolsep=2pt
\vspace{-0.4cm}
\begin{center}
\caption{Ablation study - The cross-dataset detection performance (AUC (\%)) of different mask strategies on Celeb-DF, WildDF, and DFDCP datasets.  The strategy proposed in original MAE \cite{imgmae}, cheek \& nose, and the strategy without dividing ROIs are represented as MAE strategy \cite{imgmae}, C \& N,  and w/o ROIs respectively.}
\label{masks}
\renewcommand{\tabcolsep}{3mm} 
\renewcommand{\arraystretch}{1.0}
\begin{tabular}{cccc}
\noalign{\hrule height 0.7pt}
     {Masking strategy}& {Celeb-DF}& {WildDF}&{DFDCP}\\
           
\noalign{\hrule height 0.7pt}

\textmd{MAE strategy \cite{imgmae}}&$82.1$&$78.5$&$73.7$  \\
 
\textmd{Eye}&$86.5$&$80.9$& $78.1$\\

{\textmd{C \& N}}&$85.9$&$81.0$& $76.5$ \\

{\textmd{Lip}}&$86.4$&$81.2$& $76.8$ \\

 {\textmd{w/o ROIs}}&$84.6$&$79.9$& $73.7$ \\
 
 \rowcolor{gray!20}
 {\textmd{Proposed strategy}}&${\textbf{87.1}}$&${\textbf{82.0}}$ &${\textbf{78.9}}$\\
\noalign{\hrule height 0.7pt}
 
\end{tabular}
\end{center}
\vspace{-0.7cm}
\end{table}

\noindent\textbf{Influence of the masking strategy. }We modify the masking strategy of MAE \cite{imgmae} to improve the generalization. To evaluate the effectiveness of the improved  masking strategy, we compare the proposed masking strategy with MAE's masking strategy \cite{imgmae}. Furthermore, since the modified strategy randomly selects parts to mask, evaluating the effects of  different masked parts is important. To analyze the effectiveness of the ROIs, we compare the proposed strategy with the masking strategy that does not focus on ROIs. Without changing other parts of the proposed method, we implement cross-dataset experiments with different mask strategies.

\indent The $1^{th}$ line and  $6^{th}$ line results in  Table \ref{masks} demonstrate that modifying the  masking strategy of MAE \cite{imgmae} can improve the detection performance. The results in the $2^{th}$, $3^{th}$ and $4^{th}$  lines, which represent methods that mask eye areas, cheek and nose areas, and lip areas, respectively, show a performance degradation compared to the proposed strategy.  That is, random masking a part of all facial parts is more conducive to extracting robust features than that masking a certain part only.  Moreover, the results of the $5^{th}$ line and $6^{th}$ line show that the proposed masking strategy that focuses on ROIs achieves better performance than the  masking strategy without ROIs. The reason for this is that   the model can better capture the differences between real and fake videos by masking patches in these ROIs, as fake videos typically lack 
consistency. Therefore, the proposed masking strategy illustrated in Algorithm \ref{algorithm1} is effective in detecting Deepfake videos.

\begin{table}
\setlength{\abovecaptionskip}{0.1cm}
\setlength{\belowcaptionskip}{0.1cm}
\tabcolsep=2pt
\vspace{-0.4cm}
\begin{center}
\caption{Ablation study - The cross-dataset detection performance (AUC (\%)) of without modifying MAE, without  Finetuning Network, without Mapping Network, without meta-learning, and without face reconstruction.}
\label{twostream}
\renewcommand{\tabcolsep}{2.3mm} 
\renewcommand{\arraystretch}{1.0}
\begin{tabular}{cccc}
\noalign{\hrule height 0.7pt}
     {}& {Celeb-DF}& {WildDF}&{DFDCP}\\
\noalign{\hrule height 0.7pt}
{\textmd{w/o modifying MAE}}&$75.3$& $72.1$&$70.2$ \\
 
    \textmd{w/o Finetuning Network}&$79.8$&$75.5$& $72.1$  \\
 
\textmd{w/o Mapping Network}&$79.4$& $76.5$&$76.3$\\

{\textmd{w/o Meta-learning}}&$86.7$& $81.1$&$77.6$ \\
 {\textmd{w/o face reconstruction}}&$84.9$& $80.1$&$75.4$ \\
 \rowcolor{gray!20}
{\textmd{Whole Mover}}&${\textbf{87.1}}$&${\textbf{82.0}}$ &${\textbf{78.9}}$\\
\noalign{\hrule height 0.7pt}
\end{tabular}
\end{center}
\vspace{-0.4cm}
\end{table}

\noindent\textbf{Influence of MAE.} We adopt the MAE \cite{imgmae} for Deepfake detection by modifying the masking strategy and adding the Mapping Network in the second stage. To evaluate the effectiveness of the modifications, we compared the detection performance of the modified \textsf{Mover} with the original MAE method for Deepfake detection. The results are shown in the $1^{th}$ line of Table \ref{twostream}. The detection performance of the original MAE is lower than that of \textsf{Mover}, demonstrating the effectiveness of the modifications in \textsf{Mover}. 

\noindent \textbf{Influence of Finetuning Network and Mapping Network.}  In the second stage, we combine two branches of Finetuning Network and Mapping Network for Deepfake detection. To validate the performance of each branch, we compare  the performance of a single branch with that of both branches combined. Specifically, we detect the videos by removing the Finetuning Network from \textsf{Mover}, and the results are shown in the $2^{th}$ line of Table \ref{twostream}.  We remove the Mapping Network from \textsf{Mover} to detect videos, and show the results in the $3^{th}$ line. We can see that removing either the Finetuning Network or Mapping Network degraded the detection performance, as each branch played a crucial role in Deepfake detection. Combining both branches improved the performance by magnifying the distinction between real and fake videos.

\noindent \textbf{Influence of meta-learning.} The Mapping Network uses a meta-learning module to improve the domain generalization. We also construct experiments to evaluate the effectiveness of the meta-learning module. We remove the meta-learning module to carry out experiments, and the results are shown in the $4^{th}$ line of Table \ref{twostream}. The results in the $4^{th}$ and $5^{th}$ lines show that the method without meta-learning achieves worse results than the proposed  \textsf{Mover} with meta-learning. The meta-learning approach simulates cross-domain detection in the training phase, improving cross-dataset detection performance. 

\noindent \textbf{Influence of face reconstruction.} The Mapping Network input reconstructed faces to map faces and classify Deepfake videos.  We remove the reconstructed faces to implement Mapping Network, and the results are reported in the $5^{th}$ line of Table \ref{twostream}. The results  show that the method without face reconstruction achieves worse results than the proposed  \textsf{Mover} with face reconstruction. The face reconstruction can amplify the distinctions between real and fake, resulting in improved cross-dataset detection performance. 



\begin{figure}
 \centering
\includegraphics[width=1\linewidth]{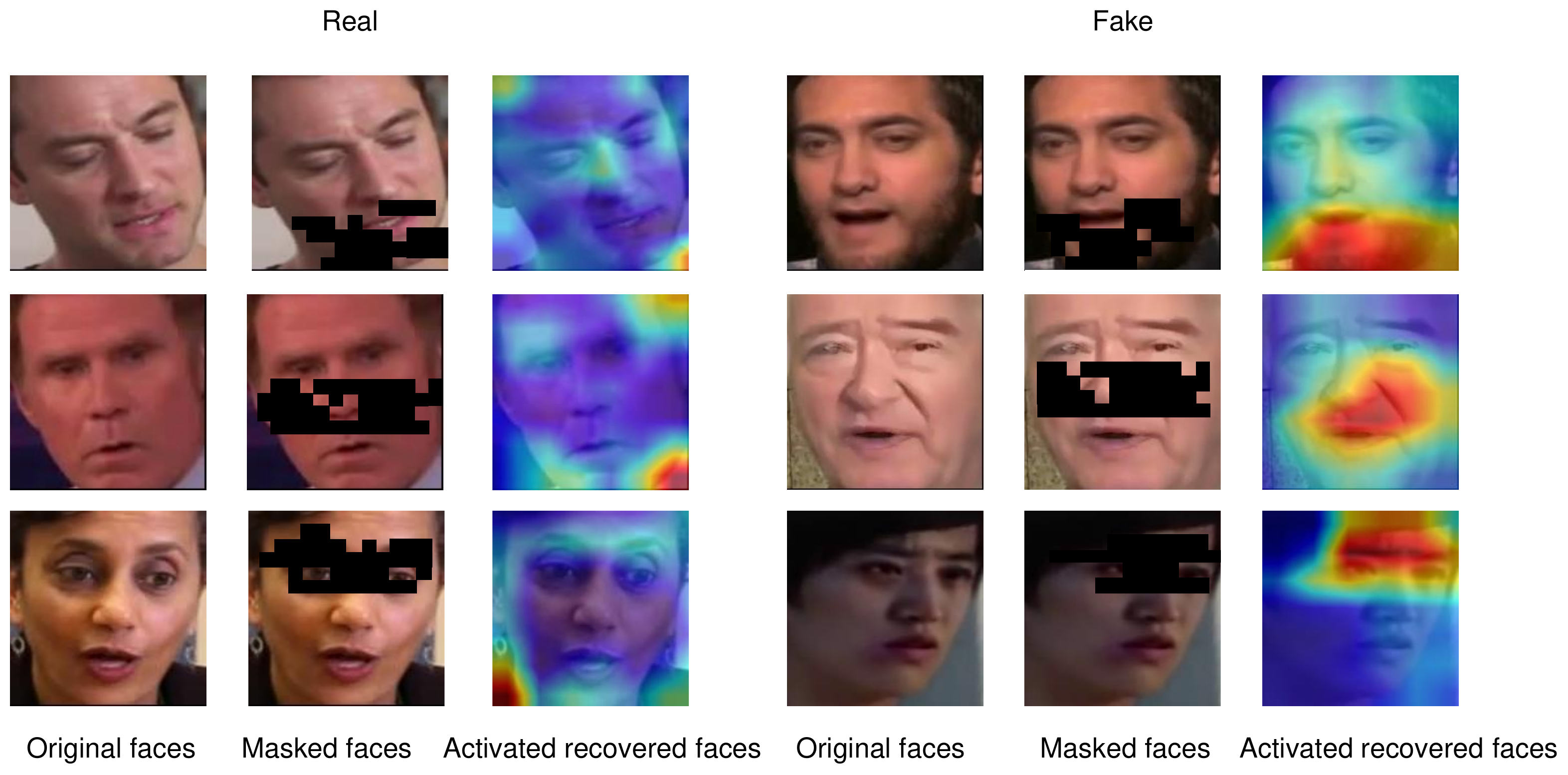}
   \vspace{-0.5cm}
\caption{The activations of \textsf{Mover}.}
 \label{fig3}
   \vspace{-0.5cm}
\end{figure}

\begin{figure}
\vspace{-0.6cm}
 \centering
\includegraphics[width=1\linewidth]{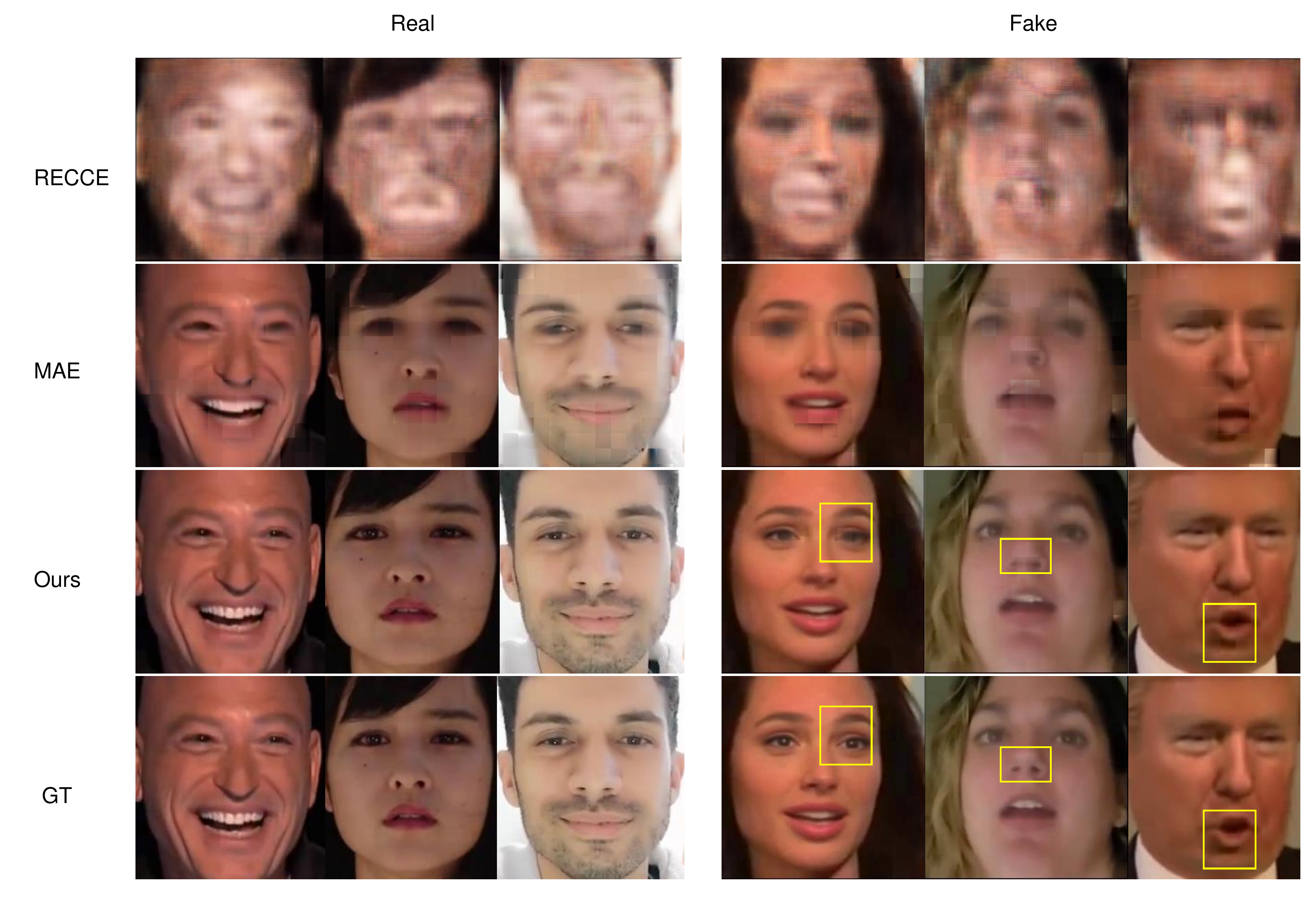}
 \vspace{-0.5cm}
\caption{The recovered faces of  RECCE\cite{recon}, MAE \cite{imgmae} and ours. RECCE\cite{recon} recovers real  and fake faces into different distributions. MAE \cite{imgmae} recovers  faces to improve
representation quality. We adopt MAE to recover faces by compelling the model to focus on facial part consistency. }
 \label{fig5}
 \vspace{-0.5cm}
\end{figure}





\noindent \textbf{Visualize the activation. } To gain insight into the features extracted by our proposed  \textsf{Mover}, 
we use the CAM function wrapped in \texttt{pytorch\_grad\_cam} to show the activations. This involves masking and reconstructing faces, and then visualizing the reconstructed faces using the trained finetuning network. As shown in Fig. \ref{fig3}, the activation areas of real and fake faces are different. For the fake face, due to the lack of consistency, the fake face is reconstructed with some perturbations. After visualizing the features of the network, the masked areas of the reconstructed fake faces are activated, exposing the inconsistencies of fake faces. In contrast, the consistency of real faces guides the reconstruction, making it difficult for the network to extract abnormal inconsistencies. Therefore, activation areas are present in the background areas rather than the masked areas. That is, the proposed \textsf{Mover} can detect videos by extracting the facial part consistencies.

\noindent\textbf{Present recovered faces. } We also present the visualization of reconstructed faces in Fig. \ref{fig5}. RECCE\cite{recon} detects the Deepfake by calculating the reconstruction difference score of  the face before and after the reconstruction and achieves promising performance, but reconstructing the whole faces without facial semantic guidance  can result in unclear reconstructions. The original MAE randomly masks images with a high masking ratio,  leading to reconstruction artifacts in both real and fake faces.  In contrast, our proposed \textsf{Mover} method randomly masks different facial parts and focuses on capturing the differences in inconsistent areas, as highlighted in the yellow box in Fig. \ref{fig5}.

\section{Conclusion}
This paper focuses on the detection of Deepfakes, particularly in identifying new types of fake videos. By focusing equally on all facial parts rather than relying on specific facial parts, our two-stage model can learn unspecific facial consistencies and robust representations that only exist on real faces.  In the first stage, the model is trained to reconstruct the entire image from partially masked ROIs on the face, which helps the model learn the facial part consistencies of real videos. In the second stage, the model is trained to maximize the differences between real and fake videos.  Extensive experiments illustrate the robustness and generalizability of \textsf{Mover} on benchmark datasets.







\begin{thebibliography}{10}\itemsep=-1pt

\bibitem{meso}
Darius Afchar, Vincent Nozick, Junichi Yamagishi, and Isao Echizen.
\newblock Mesonet: a compact facial video forgery detection network.
\newblock In {\em WIFS}, pages 1--7, 1, 2018. IEEE.

\bibitem{recon}
Junyi Cao, Chao Ma, Taiping Yao, Shen Chen, Shouhong Ding, and Xiaokang Yang.
\newblock End-to-end reconstruction-classification learning for face forgery
  detection.
\newblock In {\em CVPR}, pages 4113--4122, 1, 2022. IEEE.

\bibitem{Patch-based}
Lucy Chai, David Bau, Ser-Nam Lim, and Phillip Isola.
\newblock What makes fake images detectable? understanding properties that
  generalize.
\newblock In {\em ECCV}, pages 103--120, 1, 2020. IEEE.

\bibitem{self}
Liang Chen, Yong Zhang, Yibing Song, Lingqiao Liu, and Jue Wang.
\newblock Self-supervised learning of adversarial example: Towards good
  generalizations for deepfake detection.
\newblock In {\em CVPR}, pages 18710--18719, 1, 2022. IEEE.

\bibitem{aaailocal}
Shen Chen, Taiping Yao, Yang Chen, Shouhong Ding, Jilin Li, and Rongrong Ji.
\newblock Local relation learning for face forgery detection.
\newblock In {\em AAAI}, volume~35, pages 1081--1088, 1, 2021. IEEE.

\bibitem{chesney2019deep}
Bobby Chesney and Danielle Citron.
\newblock Deep fakes: A looming challenge for privacy, democracy, and national
  security.
\newblock {\em Calif. L. Rev.}, 107:1753, 2019.

\bibitem{fakecatcher}
Umur~Aybars Ciftci, Ilke Demir, and Lijun Yin.
\newblock Fakecatcher: Detection of synthetic portrait videos using biological
  signals.
\newblock {\em TPAMI}, 1(1), 2020.

\bibitem{deepfakegit}
DeepFakes.
\newblock Accessed october 10, 2018.
\newblock \url{https://github.com/deepfakes/faceswap}, 2018.

\bibitem{dfdcpre}
Brian Dolhansky, Russ Howes, Ben Pflaum, Nicole Baram, and Cristian~Canton
  Ferrer.
\newblock The deepfake detection challenge (dfdc) preview dataset.
\newblock {\em arXiv preprint arXiv:1910.08854}, 1(1):1, 2019.

\bibitem{dong2022protecting}
Xiaoyi Dong, Jianmin Bao, Dongdong Chen, Ting Zhang, Weiming Zhang, Nenghai Yu,
  Dong Chen, Fang Wen, and Baining Guo.
\newblock Protecting celebrities from {D}eepfake with identity consistency
  transformer.
\newblock In {\em CVPR}, pages 9468--9478, 1, 2022. IEEE.

\bibitem{vit}
Alexey Dosovitskiy, Lucas Beyer, Alexander Kolesnikov, Dirk Weissenborn,
  Xiaohua Zhai, Thomas Unterthiner, Mostafa Dehghani, Matthias Minderer, Georg
  Heigold, Sylvain Gelly, et~al.
\newblock An image is worth 16x16 words: Transformers for image recognition at
  scale.
\newblock In {\em ICLR}, page~1, 1, 2021. IEEE.

\bibitem{facs}
Paul Ekman and Wallace~V. Friesen.
\newblock Facial action coding system: a technique for the measurement of
  facial movement.
\newblock In {\em 1}, volume~1, page~1, 1, 1978. IEEE.

\bibitem{faceswapgit}
FaceSwap.
\newblock Accessed october 29, 2018.
\newblock \url{https://github.com/MarekKowalski/FaceSwap/}, 2018.

\bibitem{vimae}
Christoph Feichtenhofer, Yanghao Li, Kaiming He, et~al.
\newblock Masked autoencoders as spatiotemporal learners.
\newblock {\em NeurIPS}, 35:35946--35958, 2022.

\bibitem{conmae}
Peng Gao, Teli Ma, Hongsheng Li, Jifeng Dai, and Yu Qiao.
\newblock Convmae: Masked convolution meets masked autoencoders.
\newblock In {\em NeurIPS}, page~1, 1, 2022. IEEE.

\bibitem{stil}
Zhihao Gu, Yang Chen, Taiping Yao, Shouhong Ding, Jilin Li, Feiyue Huang, and
  Lizhuang Ma.
\newblock Spatiotemporal inconsistency learning for deepfake video detection.
\newblock In {\em ACM MM}, pages 3473--3481, 1, 2021. IEEE.

\bibitem{consis}
Zhihao Gu, Yang Chen, Taiping Yao, Shouhong Ding, Jilin Li, and Lizhuang Ma.
\newblock Delving into the local: Dynamic inconsistency learning for deepfake
  video detection.
\newblock In {\em AAAI}, volume~36, pages 744--752, 1, 2022. IEEE.

\bibitem{hcil}
Zhihao Gu, Taiping Yao, Yang Chen, Shouhong Ding, and Lizhuang Ma.
\newblock Hierarchical contrastive inconsistency learning for deepfake video
  detection.
\newblock In {\em ECCV}, pages 596--613, 1, 2022. IEEE.

\bibitem{delv}
Jiazhi Guan, Hang Zhou, Zhibin Hong, Errui Ding, Jingdong Wang, Chengbin Quan,
  and Youjian Zhao.
\newblock Delving into sequential patches for deepfake detection.
\newblock {\em NeurIPS}, 1:1, 2022.

\bibitem{haliassos2022leveraging}
Alexandros Haliassos, Rodrigo Mira, Stavros Petridis, and Maja Pantic.
\newblock Leveraging real talking faces via self-supervision for robust forgery
  detection.
\newblock In {\em CVPR}, pages 14950--14962, 1, 2022. IEEE.

\bibitem{lips}
Alexandros Haliassos, Konstantinos Vougioukas, Stavros Petridis, and Maja
  Pantic.
\newblock Lips don't lie: A generalisable and robust approach to face forgery
  detection.
\newblock In {\em CVPR}, pages 5039--5049, 1, 2021. IEEE.

\bibitem{imgmae}
Kaiming He, Xinlei Chen, Saining Xie, Yanghao Li, Piotr Doll{\'a}r, and Ross
  Girshick.
\newblock Masked autoencoders are scalable vision learners.
\newblock In {\em CVPR}, pages 16000--16009, 1, 2022. IEEE.

\bibitem{resnet}
Kaiming He, Xiangyu Zhang, Shaoqing Ren, and Jian Sun.
\newblock Deep residual learning for image recognition.
\newblock In {\em CVPR}, pages 770--778, 1, 2016. IEEE.

\bibitem{finfer}
Juan Hu, Xin Liao, Jinwen Liang, Wenbo Zhou, and Zheng Qin.
\newblock F{I}nfer: Frame inference-based deepfake detection for
  high-visual-quality videos.
\newblock In {\em AAAI}, volume~36, pages 951--959, 1, 2022. IEEE.

\bibitem{fttwostream}
Juan Hu, Xin Liao, Wei Wang, and Zheng Qin.
\newblock Detecting compressed deepfake videos in social networks using
  frame-temporality two-stream convolutional network.
\newblock {\em IEEE TCSVT}, 32(3):1089--1102, 2021.

\bibitem{meta}
Yunpei Jia, Jie Zhang, and Shiguang Shan.
\newblock Dual-branch meta-learning network with distribution alignment for
  face anti-spoofing.
\newblock {\em TIFS}, 17:138--151, 2021.

\bibitem{ffmpeg}
Xiaohua Lei, Xiuhua Jiang, and Caihong Wang.
\newblock Design and implementation of a real-time video stream analysis system
  based on ffmpeg.
\newblock In {\em WCSE}, pages 212--216, 1, 2013. IEEE.

\bibitem{li2018learning}
Da Li, Yongxin Yang, Yi-Zhe Song, and Timothy Hospedales.
\newblock Learning to generalize: Meta-learning for domain generalization.
\newblock In {\em AAAI}, volume~32, page~1, 1, 2018. IEEE.

\bibitem{frequency}
Jiaming Li, Hongtao Xie, Jiahong Li, Zhongyuan Wang, and Yongdong Zhang.
\newblock Frequency-aware discriminative feature learning supervised by
  single-center loss for face forgery detection.
\newblock In {\em CVPR}, pages 6458--6467, 1, 2021. IEEE.

\bibitem{wavelet22mm}
Jiaming Li, Hongtao Xie, Lingyun Yu, and Yongdong Zhang.
\newblock Wavelet-enhanced weakly supervised local feature learning for face
  forgery detection.
\newblock In {\em ACM MM}, pages 1299--1308, 1, 2022. IEEE.

\bibitem{xray}
Lingzhi Li, Jianmin Bao, Ting Zhang, Hao Yang, Dong Chen, Fang Wen, and Baining
  Guo.
\newblock Face x-ray for more general face forgery detection.
\newblock In {\em CVPR}, pages 5001--5010, 1, 2020. IEEE.

\bibitem{expression}
Shan Li and Weihong Deng.
\newblock Deep facial expression recognition: A survey.
\newblock {\em IEEE TAC}, 13(3):1195--1215, 2020.

\bibitem{smil}
Xiaodan Li, Yining Lang, Yuefeng Chen, Xiaofeng Mao, Yuan He, Shuhui Wang, Hui
  Xue, and Quan Lu.
\newblock Sharp multiple instance learning for deepfake video detection.
\newblock In {\em ACM MM}, pages 1864--1872, 1, 2020. IEEE.

\bibitem{eye}
Yuezun Li, Ming-Ching Chang, and Siwei Lyu.
\newblock In ictu oculi: Exposing {AI} created fake videos by detecting eye
  blinking.
\newblock In {\em WIFS}, pages 1--7, 1, 2018. IEEE.

\bibitem{fwa}
Yuezun Li and Siwei Lyu.
\newblock Exposing {D}eepfake videos by detecting face warping artifacts.
\newblock In {\em CVPRW}, pages 46--52, 1, 2019. IEEE.

\bibitem{celeb}
Yuezun Li, Xin Yang, Pu Sun, Honggang Qi, and Siwei Lyu.
\newblock Celeb-{DF}: A large-scale challenging dataset for deepfake forensics.
\newblock In {\em CVPR}, pages 3207--3216, 1, 2020. IEEE.

\bibitem{spsl}
Honggu Liu, Xiaodan Li, Wenbo Zhou, Yuefeng Chen, Yuan He, Hui Xue, Weiming
  Zhang, and Nenghai Yu.
\newblock Spatial-phase shallow learning: rethinking face forgery detection in
  frequency domain.
\newblock In {\em CVPR}, pages 772--781, 1, 2021. IEEE.

\bibitem{adamw}
Ilya Loshchilov and Frank Hutter.
\newblock Decoupled weight decay regularization.
\newblock {\em arXiv preprint arXiv:1711.05101}, 3:1, 2017.

\bibitem{luo2021generalizing}
Yuchen Luo, Yong Zhang, Junchi Yan, and Wei Liu.
\newblock Generalizing face forgery detection with high-frequency features.
\newblock In {\em CVPR}, pages 16317--16326, 1, 2021. IEEE.

\bibitem{twobranch}
Iacopo Masi, Aditya Killekar, Royston~Marian Mascarenhas, Shenoy~Pratik
  Gurudatt, and Wael AbdAlmageed.
\newblock Two-branch recurrent network for isolating deepfakes in videos.
\newblock In {\em ECCV}, pages 667--684, 1, 2020. IEEE.

\bibitem{emotions}
Trisha Mittal, Uttaran Bhattacharya, Rohan Chandra, Aniket Bera, and Dinesh
  Manocha.
\newblock Emotions don't lie: An audio-visual deepfake detection method using
  affective cues.
\newblock In {\em ACM MM}, pages 2823--2832, 1, 2020. IEEE.

\bibitem{on}
Aakash~Varma Nadimpalli and Ajita Rattani.
\newblock On improving cross-dataset generalization of deepfake detectors.
\newblock In {\em CVPR}, pages 91--99, 1, 2022. IEEE.

\bibitem{capsule}
Huy~H Nguyen, Junichi Yamagishi, and Isao Echizen.
\newblock Capsule-forensics: Using capsule networks to detect forged images and
  videos.
\newblock In {\em ICASSP}, pages 2307--2311, 1, 2019. IEEE.

\bibitem{thinking}
Yuyang Qian, Guojun Yin, Lu Sheng, Zixuan Chen, and Jing Shao.
\newblock Thinking in frequency: Face forgery detection by mining
  frequency-aware clues.
\newblock In {\em ECCV}, pages 86--103, 1, 2020. IEEE.

\bibitem{ffdata}
Andreas Rossler, Davide Cozzolino, Luisa Verdoliva, Christian Riess, Justus
  Thies, and Matthias Nie{\ss}ner.
\newblock Faceforensics++: Learning to detect manipulated facial images.
\newblock In {\em ICCV}, pages 1--11, 1, 2019. IEEE.

\bibitem{xinli}
James~A Russell and Jos{\'e} Miguel~Ed Fern{\'a}ndez-Dols.
\newblock {\em The psychology of facial expression.}
\newblock Editions de la Maison des Sciences de l'Homme, 1, 1997.

\bibitem{CNN-GRU}
Ekraam Sabir, Jiaxin Cheng, Ayush Jaiswal, Wael AbdAlmageed, Iacopo Masi, and
  Prem Natarajan.
\newblock Recurrent convolutional strategies for face manipulation detection in
  videos.
\newblock {\em Interfaces (GUI)}, 3:80--87, 2019.

\bibitem{dlib}
S Sharma, Karthikeyan Shanmugasundaram, and Sathees~Kumar Ramasamy.
\newblock Farec-cnn based efficient face recognition technique using dlib.
\newblock In {\em ICACCCT}, pages 192--195, 1, 2016. IEEE.

\bibitem{shiohara2022detecting}
Kaede Shiohara and Toshihiko Yamasaki.
\newblock Detecting {D}eepfakes with self-blended images.
\newblock In {\em CVPR}, pages 18720--18729, 1, 2022. IEEE.

\bibitem{ltw}
Ke Sun, Hong Liu, Qixiang Ye, Yue Gao, Jianzhuang Liu, Ling Shao, and Rongrong
  Ji.
\newblock Domain general face forgery detection by learning to weight.
\newblock In {\em AAAI}, volume~35, pages 2638--2646, 1, 2021. IEEE.

\bibitem{aaaidual}
Ke Sun, Taiping Yao, Shen Chen, Shouhong Ding, Jilin Li, and Rongrong Ji.
\newblock Dual contrastive learning for general face forgery detection.
\newblock In {\em AAAI}, volume~36, pages 2316--2324, 1, 2022. IEEE.

\bibitem{tan2019efficientnet}
Mingxing Tan and Quoc Le.
\newblock Efficientnet: Rethinking model scaling for convolutional neural
  networks.
\newblock In {\em ICML}, pages 6105--6114, 1, 2019. IEEE.

\bibitem{ne}
Justus Thies, Michael Zollh{\"o}fer, and Matthias Nie{\ss}ner.
\newblock Deferred neural rendering: Image synthesis using neural textures.
\newblock {\em TOG}, 38(4):1--12, 2019.

\bibitem{thies2016face2face}
Justus Thies, Michael Zollhofer, Marc Stamminger, Christian Theobalt, and
  Matthias Nie{\ss}ner.
\newblock Face2face: Real-time face capture and reenactment of rgb videos.
\newblock In {\em CVPR}, pages 2387--2395, 1, 2018. IEEE.

\bibitem{videomae}
Zhan Tong, Yibing Song, Jue Wang, and Limin Wang.
\newblock Videomae: Masked autoencoders are data-efficient learners for
  self-supervised video pre-training.
\newblock In {\em NeurIPS}, page~1, 1, 2022. IEEE.

\bibitem{mico1}
Su-Jing Wang, Wen-Jing Yan, Xiaobai Li, Guoying Zhao, and Xiaolan Fu.
\newblock Micro-expression recognition using dynamic textures on tensor
  independent color space.
\newblock In {\em ICPR}, pages 4678--4683, 1, 2014. IEEE.

\bibitem{micro2}
Su-Jing Wang, Wen-Jing Yan, Xiaobai Li, Guoying Zhao, Chun-Guang Zhou, Xiaolan
  Fu, Minghao Yang, and Jianhua Tao.
\newblock Micro-expression recognition using color spaces.
\newblock {\em IEEE Transactions on Image Processing}, 24(12):6034--6047, 2015.

\bibitem{cnn2}
Sheng-Yu Wang, Oliver Wang, Richard Zhang, Andrew Owens, and Alexei~A Efros.
\newblock Cnn-generated images are surprisingly easy to spot... for now.
\newblock In {\em CVPR}, pages 8695--8704, 1, 2020. IEEE.

\bibitem{aibase}
Zhi Wang, Yiwen Guo, and Wangmeng Zuo.
\newblock Deepfake forensics via an adversarial game.
\newblock {\em TIP}, 31:3541--3552, 2022.

\bibitem{fakefake}
Haiwei Wu, Jiantao Zhou, Jinyu Tian, Jun Liu, and Yu Qiao.
\newblock Robust image forgery detection against transmission over online
  social networks.
\newblock {\em IEEE TIFS}, 17:443--456, 2022.

\bibitem{head}
Xin Yang, Yuezun Li, and Siwei Lyu.
\newblock Exposing deep fakes using inconsistent head poses.
\newblock In {\em ICASSP}, pages 8261--8265, 1, 2019. IEEE.

\bibitem{yang2023masked}
Ziming Yang, Jian Liang, Yuting Xu, Xiao-Yu Zhang, and Ran He.
\newblock Masked relation learning for deepfake detection.
\newblock {\em TIFS}, 18:1696--1708, 2023.

\bibitem{transformer22mm}
DaiChi Zhang, Fanzhao Lin, Yingying Hua, Pengju Wang, Dan Zeng, and Shiming Ge.
\newblock Deepfake video detection with spatiotemporal dropout transformer.
\newblock In {\em ACM MM}, page 5833–5841, 1, 2022. IEEE.

\bibitem{multiatt}
Hanqing Zhao, Wenbo Zhou, Dongdong Chen, Tianyi Wei, Weiming Zhang, and Nenghai
  Yu.
\newblock Multi-attentional deepfake detection.
\newblock In {\em CVPR}, pages 2185--2194, 1, 2021. IEEE.

\bibitem{pcl}
Tianchen Zhao, Xiang Xu, Mingze Xu, Hui Ding, Yuanjun Xiong, and Wei Xia.
\newblock Learning self-consistency for deepfake detection.
\newblock In {\em ICCV}, pages 15023--15033, 1, 2021. IEEE.

\bibitem{ftcn}
Yinglin Zheng, Jianmin Bao, Dong Chen, Ming Zeng, and Fang Wen.
\newblock Exploring temporal coherence for more general video face forgery
  detection.
\newblock In {\em ICCV}, pages 15044--15054, 1, 2021. IEEE.

\bibitem{UIA-ViT}
Wanyi Zhuang, Qi Chu, Zhentao Tan, Qiankun Liu, Haojie Yuan, Changtao Miao,
  Zixiang Luo, and Nenghai Yu.
\newblock Uia-vit: Unsupervised inconsistency-aware method based on vision
  transformer for face forgery detection.
\newblock In {\em ECCV}, pages 391--407, 1, 2022. IEEE.

\bibitem{wild}
Bojia Zi, Minghao Chang, Jingjing Chen, Xingjun Ma, and Yu-Gang Jiang.
\newblock Wild{D}eepfake: A challenging real-world dataset for deepfake
  detection.
\newblock In {\em ACM MM}, pages 2382--2390, 1, 2020. IEEE.

\end{thebibliography}


\end{document}